\documentclass{article}

\usepackage[final]{nips_2017}

\usepackage[utf8]{inputenc} 
\usepackage[T1]{fontenc}    
\usepackage{hyperref}       
\usepackage{url}            
\usepackage{booktabs}       
\usepackage{amsfonts}       
\usepackage{nicefrac}       
\usepackage{microtype}      
\usepackage{graphicx}

\title{Stochastic Maximum Likelihood Optimization via Hypernetworks}

\author{
  Abdul-Saboor Sheikh, Kashif Rasul, Andreas Merentitis \& Urs Bergmann\\
  \texttt{\{saboor.sheikh, kashif.rasul, urs.bergmann\}@zalando.de} \\ 
  \texttt{andreas.merentitis@ieee.org}\\
  Zalando Research, Zalando SE\\
  Mühlenstraße 25, 10243 Berlin, Germany
}

\begin{document}

\maketitle

\begin{abstract}
  This work explores maximum likelihood optimization of neural networks through hypernetworks. A hypernetwork initializes the weights of another network, which in turn can be employed for typical functional tasks such as regression and classification. We optimize hypernetworks to directly maximize the conditional likelihood of target variables given input. Using this approach  we obtain competitive empirical results on regression and classification benchmarks.
\end{abstract}

\section{Introduction}

Implicit distribution (ID) representation has lately been a subject of much interest. IDs govern variables that result from linear or non-linear transformations applied to another set of (standard) random variates. Through successive transformations, IDs can be equipped with enough capacity to represent functions of arbitrary complexity (i.e., high-dimensionality, multimodality, etc.). This has seen IDs become mainstream in approximate probabilistic inference in deep learning, where they have been extensively applied to approximate intractable posterior distributions over latent variables and model parameters conditioned on data~\citep[][among many others]{DBLP:journals/corr/KingmaW13, DBLP:conf/icml/RezendeM15, DBLP:conf/nips/LiuW16, DBLP:journals/corr/WangL16f, DBLP:conf/nips/RanganathTAB16, Mescheder2017ICML, DBLP:journals/corr/TranRB17,DBLP:journals/corr/Huszar17}.
Although the probability density function of an ID may not be accessible (e.g., due to non-invertible mappings), the reparameterization trick~\cite{DBLP:journals/corr/KingmaW13} still allows for an efficient and unbiased means for sampling it. This makes IDs amenable to (gradient-based) stochastic optimization regimes.

\begin{figure}
\begin{center}
\includegraphics[width=0.5\textwidth]{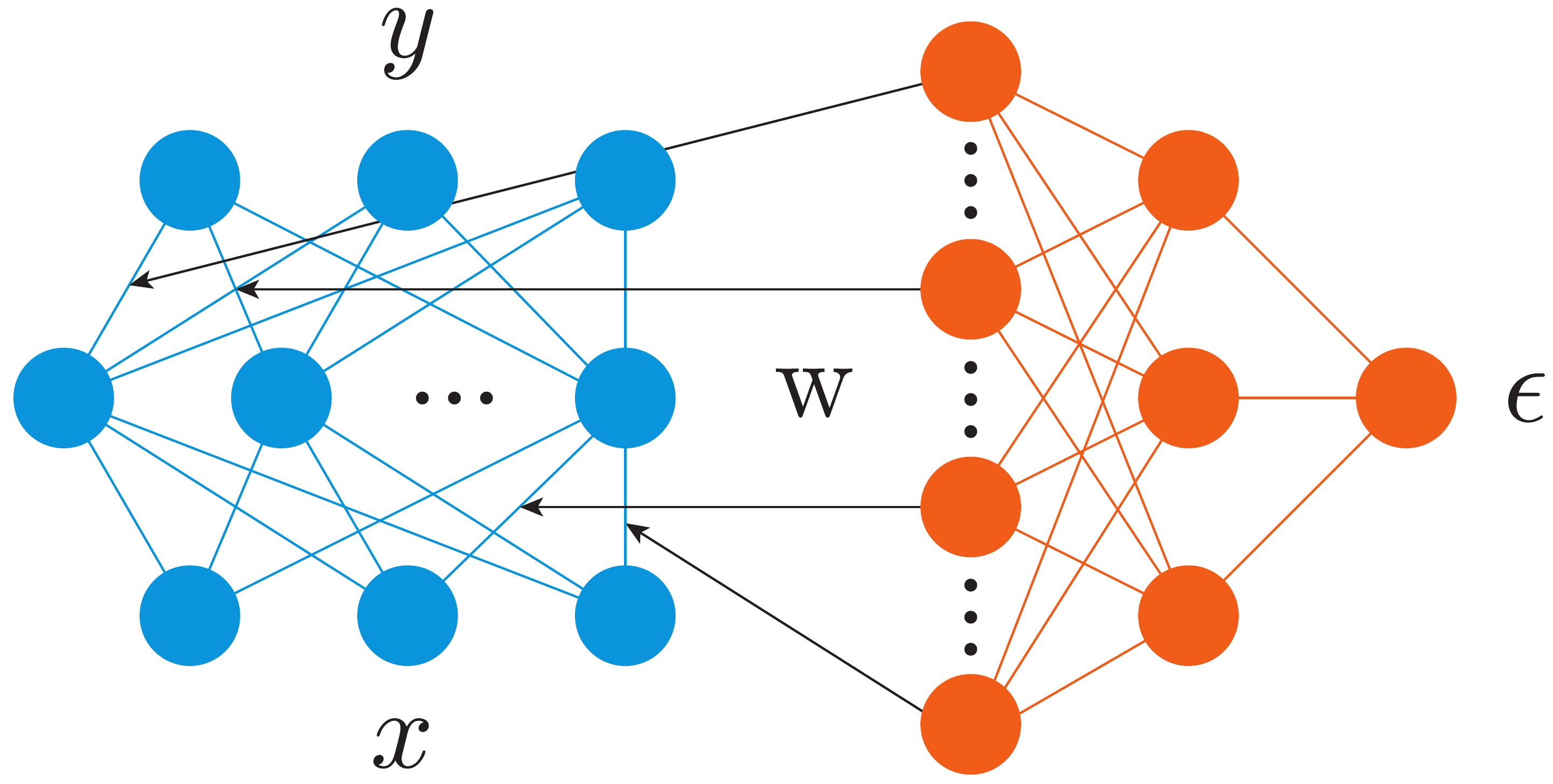}
\end{center}
\caption{Schematic of a hypernetwork that takes a noise variable $\epsilon$ and outputs the weights of the main network.}
\label{figure:hypernet}
\end{figure}

In this work we employ IDs as hypernetworks. The idea behind a \emph{hypernetwork} (Figure \ref{figure:hypernet}) is to have a neural network that outputs the weights of another \emph{main} neural network. The main network is then applied for standard tasks such as regression or classification. We optimize the parameters of the hypernetwork using stochastic gradient descent (SGD e.g.~\cite{DBLP:journals/corr/KingmaB14}) to maximize the (marginalized) conditional likelihood of targets given input. In empirical analysis on various regression and classification benchmarks, we find our approach to be in general better than standard maximum likelihood learning, while it also performs competitively with respect to a number of approximate Bayesian deep learning methodologies.

\section{Related Work}

Our optimization framework is not fully Bayesian; however instead of maintaining point parameter estimates as in standard maximum likelihood optimization of neural networks, our approach optimizes an ID that governs the parameters of a neural network. In contrast to Bayesian Neural Networks (BNNs)~\citep[][among others]{Neal:1996:BLN:525544, MACKAY199573, ICML-2011-WellingT,NIPS2011_4329,pmlr-v37-hernandez-lobatoc15,pmlr-v70-li17a}, where the target is to approximate the posterior distribution of neural network parameters (e.g., using variational inference~\cite{NIPS2011_4329}), we deploy and tune the parameter distribution while directly optimizing for the objective of the main neural network. Hence our approach is much akin to gradient-based evolutionary optimization~\cite{DBLP:journals/corr/SunWS12}.

\subsection{Hypernetworks}

 This work uses the hypernetworks approach, which has its roots in this paper~\cite{Schmidhuber:1992:LCF:148061.148068}. The term itself was introduced in the paper~\cite{2016arXiv160909106H} where the complete model was trained via backpropagation, like in this paper, and used efficiently compressed weights to reduce the total size of the network.

\subsection{Bayesian Neural Networks}

Studied since the mid 90's \cite{Neal:1996:BLN:525544} \cite{MACKAY199573} BNNs can be classified into two main methods, either using MCMC~\cite{ICML-2011-WellingT} or learning an approximate posterior using stochastic variational inference (VI)~\cite{NIPS2011_4329}, expectation propagation~\cite{pmlr-v37-hernandez-lobatoc15} or $\alpha$-divergence~\cite{pmlr-v70-li17a}. In the VI setting one can interpret dropout~\cite{2015arXiv150602142G} as a VI method allowing cheap samples from $q(\theta)$ but resulting in a unimodal approximate posterior. Bayes by Backprop~\cite{Blundell:2015:WUN:3045118.3045290} can also be viewed as a simple Bayesian hypernet, where the hypernetwork only performs element wise shift and scale of the noise. To solve the issue of scaling~\cite{2017arXiv170301961L} propose a hypernet that generates scaling factors on the means of a factorial Gaussian distribution allowing for a highly flexible approximate posterior, but requiring an auxiliary inference network for the entropy term of the variational lower bound. Finally,~\cite{2017arXiv171004759K}  is closely related to this work, where a hypernetwork  learns to transform a noise distribution to a distribution of weights of a target network, and is trained via VI.

\section{Method}

Given training data $\mathcal{D}$ containing $(X,Y) = \{x^{(n)}, y^{(n)}\}_{n = 1}^{N}$, we consider the task of maximizing the marginal conditional likelihood of targets $Y$ given $X$:
\begin{equation}
\label{eqn:condLkh}
  p(Y|X)  = \int \prod_n p(y^{(n)} | x^{(n)},\theta)\, p(\theta)\, \mathrm{d}\theta  
  		\ \approx \ \frac{1}{S} \sum_s \prod_n p(y^{(n)} | x^{(n)},\theta_s); \quad  \theta_s \sim p(\theta), 
\end{equation}
where the approximation allows us to estimate the intractable integral by Monte Carlo sampling. While  the logarithm of $ p(y | x,\theta)$ in (\ref{eqn:condLkh}) can be taken to be proportional to a loss function $l(y,f(x;\mathrm{w})) := \parallel y - f(x;\mathrm{w}) \parallel$ defined between the ground-truth $y$ and the output of a deep neural network $f(x;\mathrm{w})$, $p(\theta)$ (with $\theta = \mathrm{w}$) can be chosen to have any parametric form, such as fully factorized Gaussian distributions across each dimension of the parameter vector $\mathrm{w}$. Such non-structured distributions would however be undesired for not allowing dependencies among dimensions of $\mathrm{w}$. Also from a sampling point of view it would be inefficient, in particular in high-dimensional cases. We therefore define an implicit distribution $q(\mathrm{w}; \Theta)$ such that it has a capacity to model arbitrary dependency structures among $\mathrm{w}$, while by means of the reparameterization trick~\cite{DBLP:journals/corr/KingmaW13}, it also allows for cheap sampling $\mathrm{w} \sim q(\mathrm{w}; \Theta)$ by constructing $\mathrm{w}$ as:
\begin{equation}
	\mathrm{w} =  g(\epsilon;\Theta); \quad \epsilon \sim p(\epsilon), 
\label{eqn:hypNwk}
\end{equation}
where $g(\epsilon;\Theta)$ is a hypernetwork that gets activated by a latent noise variable  $\epsilon$ drawn from a standard distribution $p(\epsilon)$, such as a Gaussian or uniform distribution. Given (\ref{eqn:hypNwk}) and the main network $f(x;\mathrm{w})$, we can write down the following function as an approximation to the logarithm of~(\ref{eqn:condLkh}):
\begin{equation}
	\mathcal{F}(\mathcal{D},\Theta) =   \sum_n l(y^{(n)}, f(x^{(n)}; \mathrm{w}_s)); \quad \mathrm{w}_s \sim q(\mathrm{w}; \Theta).
\label{eqn:objFnc}
\end{equation}

Since we draw one parameter sample per data point, we have dropped the summation over $S$ in (\ref{eqn:objFnc}). Given $\mathcal{D}$ we can optimize (\ref{eqn:objFnc}) w.r.t. $\Theta$ by applying gradient-based optimization schemes such as SGD. To predict $y_{*}$ given an input $x_{*}$ we have:

\begin{equation}
	y_{*} = \int f(x_{*}; \mathrm{w})\, q(\mathrm{w};\Theta)\, \mathrm{d}\mathrm{w} \ \approx \  
    \frac{1}{S} \sum_s f(x_{*}; \mathrm{w}_s);\ \ \ \mathrm{w}_s \sim q(\mathrm{w};\Theta). \nonumber
\end{equation}

The objective function (\ref{eqn:objFnc}) is a stochastic relaxation of standard maximum likelihood optimization. It encourages the hypernetwork to mainly reproduce outputs for which the conditional likelihood was found to be maximal. This implies that $q(\mathrm{w};\Theta)$ may eventually either converge to a delta peak at a (local) optimum or have its mass distributed across equally (sub)optimal regions, which in principle can coincide with the solution found by plain maximum likelihood (i.e., standard gradient descent techniques). In practice we however observe that our approach finds solutions that are generally better than standard gradient descent and are on par with more sophisticated Bayesian deep learning methodologies (see Sec.~\ref{sec:experiments} for more details).
This may be due to the fact that we start with a large population of solutions that shrinks down gradually as $q(\mathrm{w};\Theta)$ continues to concentrate its mass to the region(s) of most promising solutions seen so far. This can be seen as akin to gradient-based efficient evolution strategies~\cite{DBLP:journals/corr/SunWS12}.

\section{Experiments}
\label{sec:experiments}
We validate our approach in accordance with relevant literature on standard regression tasks on UCI datasets as well as MNIST classification. We use ReLU non-linearity in all our experiments for hidden layers in both hyper and main networks. We use dropout probability of 0.5 for the hidden units of hypernetworks during training. We use a softplus layer to scale (initially $1.0$) the input to the hypernetwork in (\ref{eqn:hypNwk}). For both our approach and standard SGD results reported below, we add Gaussian noise to input $x$ in (\ref{eqn:objFnc}), with its scale (initially $0.5$) tuned during end to end training as the output of a softplus unit.

For the regression experiments we follow the process as in~\cite{pmlr-v37-hernandez-lobatoc15}: we randomly keep 10\% of the data for testing and the rest for training.
Following the earlier works~\cite{pmlr-v37-hernandez-lobatoc15, Louizos:2016:SEV:3045390.3045571}, the main network architecture consists of a single hidden layer with 50 units (or 100 for the Protein and Year).  
For datasets with more than 5,000 data points (Kin8nm, Naval, Power, Protein and Year) we switch from mini-batches of size 1 to 512 (not fine-tuned). We run a total of 2000 epochs per repetition. As this number is not fine-tuned, it could cause overfitting in smaller datasets like Boston. Early stopping in such cases might improve the results.

\begin{figure}
\begin{center}
\includegraphics[width=\textwidth]{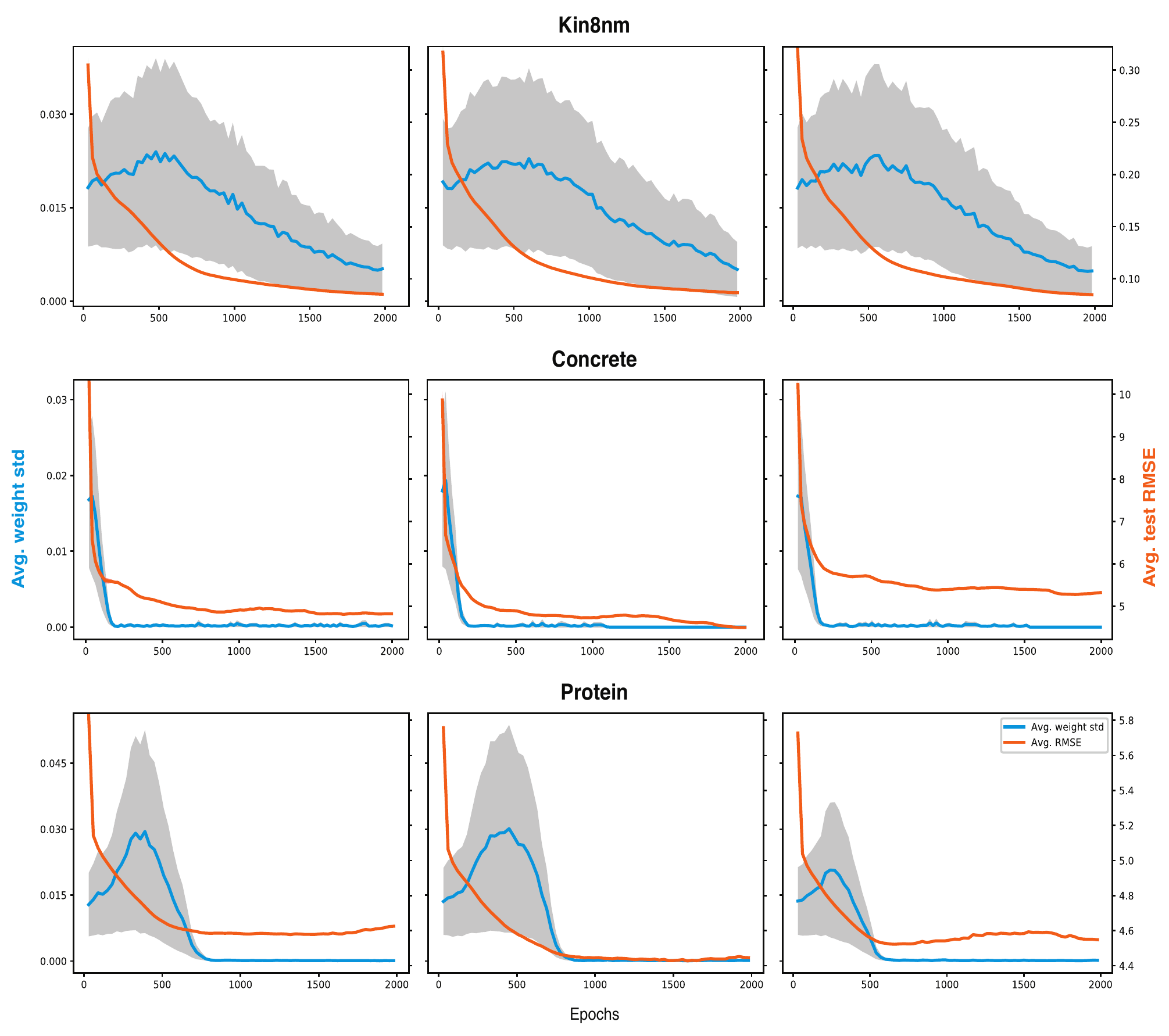} 
\end{center}
\caption{Each row plots three random repetitions of the regression task over a benchmark. The blue curves are the means of the standard deviations computed across each dimension of the hypernetwork output. The values are computed at every epoch from $10$ samples of the hypernetwork output (i.e., $10$ initializations of the main network). The gray areas cover $3$ standard deviations around the mean values. The red curves are the average test set RMSE computed per epoch as for our results in Table~\ref{table:rmse} .}
\label{figure:weight_std_vs_rmse}
\end{figure}

We compare the results of our approach with the results of other state of the art methods, as well as plain SGD trained on the same target network $f(x;w)$ without employing a hypernetwork. For each task we output the mean RMSE over the last 20 epochs and then repeat the task 20 times to report the mean of these RMSEs and their standard errors in Table~\ref{table:rmse} for our method and SGD. It can be seen that in most cases the results are competitive, and often exceed the state of the art. In Figure~\ref{figure:weight_std_vs_rmse} we further plot how the distribution over weights as represented by the hypernetwork evolves over training epochs for three (out of $20$) independent repetitions of three of the regression benchmarks. While we observe a consistent diffusion behavior across multiple repetitions on a given dataset, we see that for different datasets the diffusion trajectories leading to convergence (to a delta peak) can vary notably. In the plots we also overlay the evolution of the average test RMSE for a comparison.

\begin{table}[htb]
\centering
\resizebox{\textwidth}{!}{%
\begin{tabular}{lccccccccc}
\textbf{Dataset} & $N$ & $d$ & VI            & PBP           & Dropout                & VMG         & SGD   & Our \\ \hline
Boston & 506 & 13          & 4.32$\pm$0.29 & 3.01$\pm$0.18 & 2.97$\pm$0.85          & \textbf{2.70$\pm$0.13} &  3.73 $\pm$ 0.67 & 3.72 $\pm$ 1.05\\
Concrete  & 1,030 & 8       & 7.19$\pm$0.12 & 5.67$\pm$0.09 & 5.23$\pm$ 0.53         & 4.89$\pm$0.12 &  5.29 $\pm$   0.87 & \textbf{4.74 $\pm$  0.64}   \\
Energy    & 768 & 8       & 2.65$\pm$0.08 & 1.80$\pm$0.05 & 1.66$\pm$0.19          & \textbf{0.54$\pm$0.02} & 0.95 $\pm$ 0.13  & 0.87 $\pm$ 0.10 \\
Kin8nm    & 8,192 & 8       & 0.10$\pm$0.00 & 0.10$\pm$0.00 & 0.10$\pm$0.00          & \textbf{0.08$\pm$0.00} & \textbf{0.08 $\pm$ 0.00}   & \textbf{0.08 $\pm$ 0.00} \\
Naval & 11,934 & 16           & 0.01$\pm$0.00 & 0.01$\pm$0.00 & 0.01$\pm$0.00          & \textbf{0.00$\pm$0.00} & \textbf{0.00 $\pm$ 0.00}  & \textbf{0.00 $\pm$ 0.00}  \\
Power Plant & 9,568 & 4      & 4.33$\pm$0.04 & 4.12$\pm$0.03 & \textbf{4.02$\pm$0.18} & 4.04$\pm$0.04   &   4.06 $\pm$ 0.25   & \textbf{4.02 $\pm$ 0.18} \\
Protein  & 45,730 & 9        & 4.84$\pm$0.03 & 4.73$\pm$0.01 & 4.36$\pm$0.04          & \textbf{4.13$\pm$0.02} & 4.37 $\pm$ 0.03  & 4.65 $\pm$  0.19\\
Wine   & 1,599 & 11          & 0.65$\pm$0.01 & 0.64$\pm$0.01 & 0.62$\pm$0.04 & 0.63$\pm$0.01   &   0.80 $\pm$ 0.05   & \textbf{0.62 $\pm$ 0.04} \\
Yacht   & 308 & 6         & 6.89$\pm$0.67 & 1.02$\pm$0.05 & 1.11$\pm$0.38          & 0.71$\pm$0.05 &   0.77 $\pm$ 0.25 & \textbf{0.57 $\pm$ 0.21} \\
Year   & 515,345 & 90          & 9.03$\pm$NA  & 8.87$\pm$NA  & 8.84$\pm$NA           & 8.78$\pm$NA  &  \textbf{8.74 $\pm$ 0.03}  & \textbf{8.74 $\pm$ 0.03}   \\ \hline
\end{tabular}%
}
\caption{Average test set RMSE and standard errors for the regression datasets for VI method of~\cite{NIPS2011_4329}, PBP method of~\cite{pmlr-v37-hernandez-lobatoc15}, Dropout uncertainty of~\cite{2015arXiv150602142G}, VMG of~\cite{Louizos:2016:SEV:3045390.3045571}, standard  SGD and our method.}
\label{table:rmse}
\end{table}

For the classification task, we used  MNIST to train with varying numbers of layers and hidden units per layer, as in Table~\ref{table:mnist} for 2,000 epochs. We use a mini-batch size of 256. The hypernetwork for the 2$\times$400 and 2$\times$800  cases consist of 2 hidden layers of size 100 and 50 and for the 3$\times$150 we have 2 hidden layers of 100 and 20.
The error rate reported is calculated by taking the means of test errors from the last 20 epochs for our method and plain SGD.

\begin{table}[htb]
\centering
\begin{tabular}{lccccccc}
\textbf{Arch.} & Max. Likel.  & DropConnect & Bayes B. SM & Var. Dropout & VMG & SGD & Our \\ \hline
2$\times$400 &  --  & -- & 1.36 & -- & \textbf{1.15} & 1.43 & 1.26 \\
2$\times$800 & 1.60  & 1.20 & 1.34 & -- & -- & 1.64 & 1.37 \\
3$\times$150 &  --  & -- & -- & $\approx$ 1.42 & 1.18 & 1.71 & 1.49 \\
3$\times$750 & -- & -- & -- & $\approx$ 1.09 & \textbf{1.05} & -- & -- \\ \hline
\end{tabular}
\caption{Test errors for the MNIST dataset for Max. Likel.~\cite{Simard:2003:BPC:938980.939477}, DropConnect~\cite{icml2013_wan13}, Bayes by B. with scale mixture~\cite{Blundell:2015:WUN:3045118.3045290}, Var. Dropout~\cite{NIPS2015_5666}, VMG~\cite{Louizos:2016:SEV:3045390.3045571}, mean for plain SGD and mean for our method.}
\label{table:mnist}
\end{table}

\section{Conclusion}
We have presented a simple stochastic optimization approach for deep neural networks. The method employs implicit distributions as hypernetworks to model arbitrary dependencies among parameters of the main network. Despite being not fully Bayesian, our approach aims to model a distribution over the parameters of a neural network, as opposed to maintaining point weight estimates as in standard maximum likelihood optimization of neural networks. In general we empirically outperform standard gradient descent optimization and demonstrate on par performance in a broader comparison with state of the art Bayesian methodologies on regression and classification tasks. 

In the future we would like to focus on the scalability of our approach (e.g., through layer coupling~\cite{2017arXiv171004759K}) as well as on a fully Bayesian extension of our optimization procedure.

\clearpage
\bibliography{references}{}

\begin{thebibliography}{}

\bibitem[\protect\astroncite{Blundell
  et~al.}{2015}]{Blundell:2015:WUN:3045118.3045290}
Blundell, C., J.~Cornebise, K.~Kavukcuoglu, and
  D.~Wierstra\leavevmode\nopagebreak\newline 2015.
\newblock Weight uncertainty in neural networks.
\newblock In {\em Proceedings of the 32Nd International Conference on
  International Conference on Machine Learning - Volume 37}, ICML'15, Pp.~
  1613--1622. JMLR.org.

\bibitem[\protect\astroncite{{Gal} and
  {Ghahramani}}{2015}]{2015arXiv150602142G}
{Gal}, Y. and Z.~{Ghahramani}\leavevmode\nopagebreak\newline 2015.
\newblock {Dropout as a Bayesian Approximation: Representing Model Uncertainty
  in Deep Learning}.
\newblock {\em ArXiv e-prints}.

\bibitem[\protect\astroncite{Graves}{2011}]{NIPS2011_4329}
Graves, A.\leavevmode\nopagebreak\newline 2011.
\newblock Practical variational inference for neural networks.
\newblock In {\em Advances in Neural Information Processing Systems 24},
  J.~Shawe-Taylor, R.~S. Zemel, P.~L. Bartlett, F.~Pereira, and K.~Q.
  Weinberger, eds., Pp.~ 2348--2356.
\newblock Curran Associates, Inc.

\bibitem[\protect\astroncite{{Ha} et~al.}{2016}]{2016arXiv160909106H}
{Ha}, D., A.~{Dai}, and Q.~V. {Le}\leavevmode\nopagebreak\newline 2016.
\newblock {HyperNetworks}.
\newblock {\em ArXiv e-prints}.

\bibitem[\protect\astroncite{Hernandez-Lobato and
  Adams}{2015}]{pmlr-v37-hernandez-lobatoc15}
Hernandez-Lobato, J.~M. and R.~Adams\leavevmode\nopagebreak\newline 2015.
\newblock Probabilistic backpropagation for scalable learning of bayesian
  neural networks.
\newblock In {\em Proceedings of the 32nd International Conference on Machine
  Learning}, F.~Bach and D.~Blei, eds., volume~37 of {\em Proceedings of
  Machine Learning Research}, Pp.~ 1861--1869, Lille, France. PMLR.

\bibitem[\protect\astroncite{Husz{\'{a}}r}{2017}]{DBLP:journals/corr/Huszar17}
Husz{\'{a}}r, F.\leavevmode\nopagebreak\newline 2017.
\newblock Variational inference using implicit distributions.
\newblock {\em CoRR}, abs/1702.08235.

\bibitem[\protect\astroncite{Kingma and
  Ba}{2014}]{DBLP:journals/corr/KingmaB14}
Kingma, D.~P. and J.~Ba\leavevmode\nopagebreak\newline 2014.
\newblock Adam: {A} method for stochastic optimization.
\newblock {\em CoRR}, abs/1412.6980.

\bibitem[\protect\astroncite{Kingma et~al.}{2015}]{NIPS2015_5666}
Kingma, D.~P., T.~Salimans, and M.~Welling\leavevmode\nopagebreak\newline 2015.
\newblock Variational dropout and the local reparameterization trick.
\newblock In {\em Advances in Neural Information Processing Systems 28},
  C.~Cortes, N.~D. Lawrence, D.~D. Lee, M.~Sugiyama, and R.~Garnett, eds., Pp.~
  2575--2583.
\newblock Curran Associates, Inc.

\bibitem[\protect\astroncite{Kingma and
  Welling}{2013}]{DBLP:journals/corr/KingmaW13}
Kingma, D.~P. and M.~Welling\leavevmode\nopagebreak\newline 2013.
\newblock Auto-encoding variational bayes.
\newblock {\em CoRR}, abs/1312.6114.

\bibitem[\protect\astroncite{{Krueger} et~al.}{2017}]{2017arXiv171004759K}
{Krueger}, D., C.-W. {Huang}, R.~{Islam}, R.~{Turner}, A.~{Lacoste}, and
  A.~{Courville}\leavevmode\nopagebreak\newline 2017.
\newblock {Bayesian Hypernetworks}.
\newblock {\em ArXiv e-prints}.

\bibitem[\protect\astroncite{Li and Gal}{2017}]{pmlr-v70-li17a}
Li, Y. and Y.~Gal\leavevmode\nopagebreak\newline 2017.
\newblock Dropout inference in {B}ayesian neural networks with
  alpha-divergences.
\newblock In {\em Proceedings of the 34th International Conference on Machine
  Learning}, D.~Precup and Y.~W. Teh, eds., volume~70 of {\em Proceedings of
  Machine Learning Research}, Pp.~ 2052--2061, International Convention Centre,
  Sydney, Australia. PMLR.

\bibitem[\protect\astroncite{Liu and Wang}{2016}]{DBLP:conf/nips/LiuW16}
Liu, Q. and D.~Wang\leavevmode\nopagebreak\newline 2016.
\newblock Stein variational gradient descent: {A} general purpose bayesian
  inference algorithm.
\newblock In {\em Advances in Neural Information Processing Systems 29: Annual
  Conference on Neural Information Processing Systems 2016, December 5-10,
  2016, Barcelona, Spain}, Pp.~ 2370--2378.

\bibitem[\protect\astroncite{Louizos and
  Welling}{2016}]{Louizos:2016:SEV:3045390.3045571}
Louizos, C. and M.~Welling\leavevmode\nopagebreak\newline 2016.
\newblock Structured and efficient variational deep learning with matrix
  gaussian posteriors.
\newblock In {\em Proceedings of the 33rd International Conference on
  International Conference on Machine Learning - Volume 48}, ICML'16, Pp.~
  1708--1716. JMLR.org.

\bibitem[\protect\astroncite{{Louizos} and
  {Welling}}{2017}]{2017arXiv170301961L}
{Louizos}, C. and M.~{Welling}\leavevmode\nopagebreak\newline 2017.
\newblock {Multiplicative Normalizing Flows for Variational Bayesian Neural
  Networks}.
\newblock {\em ArXiv e-prints}.

\bibitem[\protect\astroncite{MacKay}{1995}]{MACKAY199573}
MacKay, D.~J.\leavevmode\nopagebreak\newline 1995.
\newblock Bayesian neural networks and density networks.
\newblock {\em Nuclear Instruments and Methods in Physics Research Section A:
  Accelerators, Spectrometers, Detectors and Associated Equipment}, 354(1):73
  -- 80.
\newblock Proceedings of the Third Workshop on Neutron Scattering Data
  Analysis.

\bibitem[\protect\astroncite{Mescheder et~al.}{2017}]{Mescheder2017ICML}
Mescheder, L., S.~Nowozin, and A.~Geiger\leavevmode\nopagebreak\newline 2017.
\newblock Adversarial variational bayes: Unifying variational autoencoders and
  generative adversarial networks.
\newblock In {\em International Conference on Machine Learning (ICML) 2017}.

\bibitem[\protect\astroncite{Neal}{1996}]{Neal:1996:BLN:525544}
Neal, R.~M.\leavevmode\nopagebreak\newline 1996.
\newblock {\em Bayesian Learning for Neural Networks}.
\newblock Secaucus, NJ, USA: Springer-Verlag New York, Inc.

\bibitem[\protect\astroncite{Ranganath
  et~al.}{2016}]{DBLP:conf/nips/RanganathTAB16}
Ranganath, R., D.~Tran, J.~Altosaar, and D.~M.
  Blei\leavevmode\nopagebreak\newline 2016.
\newblock Operator variational inference.
\newblock In {\em Advances in Neural Information Processing Systems 29: Annual
  Conference on Neural Information Processing Systems 2016, December 5-10,
  2016, Barcelona, Spain}, Pp.~ 496--504.

\bibitem[\protect\astroncite{Rezende and
  Mohamed}{2015}]{DBLP:conf/icml/RezendeM15}
Rezende, D.~J. and S.~Mohamed\leavevmode\nopagebreak\newline 2015.
\newblock Variational inference with normalizing flows.
\newblock In {\em Proceedings of the 32nd International Conference on Machine
  Learning, {ICML} 2015, Lille, France, 6-11 July 2015}, Pp.~ 1530--1538.

\bibitem[\protect\astroncite{Schmidhuber}{1992}]{Schmidhuber:1992:LCF:148061.148068}
Schmidhuber, J.\leavevmode\nopagebreak\newline 1992.
\newblock Learning to control fast-weight memories: An alternative to dynamic
  recurrent networks.
\newblock {\em Neural Comput.}, 4(1):131--139.

\bibitem[\protect\astroncite{Simard
  et~al.}{2003}]{Simard:2003:BPC:938980.939477}
Simard, P.~Y., D.~Steinkraus, and J.~C. Platt\leavevmode\nopagebreak\newline
  2003.
\newblock Best practices for convolutional neural networks applied to visual
  document analysis.
\newblock In {\em Proceedings of the Seventh International Conference on
  Document Analysis and Recognition - Volume 2}, ICDAR '03, Pp.~ 958--,
  Washington, DC, USA. IEEE Computer Society.

\bibitem[\protect\astroncite{Sun et~al.}{2012}]{DBLP:journals/corr/SunWS12}
Sun, Y., D.~Wierstra, T.~Schaul, and
  J.~Schmidhuber\leavevmode\nopagebreak\newline 2012.
\newblock Efficient natural evolution strategies.
\newblock {\em CoRR}, abs/1209.5853.

\bibitem[\protect\astroncite{Tran et~al.}{2017}]{DBLP:journals/corr/TranRB17}
Tran, D., R.~Ranganath, and D.~M. Blei\leavevmode\nopagebreak\newline 2017.
\newblock Deep and hierarchical implicit models.
\newblock {\em CoRR}, abs/1702.08896.

\bibitem[\protect\astroncite{Wan et~al.}{2013}]{icml2013_wan13}
Wan, L., M.~Zeiler, S.~Zhang, Y.~L. Cun, and
  R.~Fergus\leavevmode\nopagebreak\newline 2013.
\newblock Regularization of neural networks using dropconnect.
\newblock In {\em Proceedings of the 30th International Conference on Machine
  Learning (ICML-13)}, S.~Dasgupta and D.~Mcallester, eds., volume~28, Pp.~
  1058--1066. JMLR Workshop and Conference Proceedings.

\bibitem[\protect\astroncite{Wang and Liu}{2016}]{DBLP:journals/corr/WangL16f}
Wang, D. and Q.~Liu\leavevmode\nopagebreak\newline 2016.
\newblock Learning to draw samples: With application to amortized {MLE} for
  generative adversarial learning.
\newblock {\em CoRR}, abs/1611.01722.

\bibitem[\protect\astroncite{Welling and Teh}{2011}]{ICML-2011-WellingT}
Welling, M. and Y.~W. Teh\leavevmode\nopagebreak\newline 2011.
\newblock {Bayesian Learning via Stochastic Gradient Langevin Dynamics}.
\newblock In {\em {Proceedings of the 28th International Conference on Machine
  Learning}}, Pp.~ 681--688. {Omnipress}.

\end{thebibliography}
\bibliographystyle{plain}
\end{document}